# Domain Specific Approximation for Object Detection

Ting-Wu Chin[1], Chia-Lin Yu[1], Matthew Halpern[2], Hasan Genc[2], Shiao-Li Tsao[1], and Vijay Janapa Reddi[2]

*Abstract*—There is growing interest in object detection in advanced driver assistance systems and autonomous robots and vehicles. To enable such innovative systems, we need faster object detection. In this work, we investigate the trade-off between accuracy and speed with domain-specific approximations, i.e. category-aware image size scaling and proposals scaling, for two state-of-the-art deep learning-based object detection meta-architectures. We study the effectiveness of applying approximation both statically and dynamically to understand the potential and the applicability of them. By conducting experiments on the ImageNet VID dataset, we show that domain-specific approximation has great potential to improve the speed of the system without deteriorating the accuracy of object detectors, i.e. up to 7.5x speedup for dynamic domain-specific approximation. To this end, we present our insights toward harvesting domain-specific approximation as well as devise a proof-of-concept runtime, AutoFocus, that exploits dynamic domain-specific approximation.

## I. Introduction

With rapid progress being made in the field of computer vision and machine learning, there is growing interests in the practical deployment of the intelligent algorithms in systems, such as autonomous vehicles and robots, and advanced driver assistance systems. For many of these systems, detection is one of the fundamental algorithms involved in developing end- to- end applications. Detection can lead to obstacle recognition, avoidance, and navigation. As a result, detection is becoming an important algorithm for developing cognitive visual agents.

Thus far, the majority of effort on object detection has been focused on achieving high accuracy. However, from a system's perspective, object detection speed also matters. For instance, how fast an autonomous car or drone can move depends on how fast the detection algorithms work. Moreover, to build robust, intelligent agents, there will likely be more than one cognitive algorithm running on the system at the same time.

Therefore, the throughput of the system ought to be one of the top concerns for system designers in the near future. In our experiments, modern deep learning-based object detectors, without modification, could only provide up to 2 Frames-per-second (FPS) on NVidia's Tegra X1 with 640x480 images. Typical real-time performance requires about 30 FPS. We need faster hardware to ensure real-time performance while several tasks are running simultaneously and to cope with future high resolution (e.g. 2048x1024) images. Alternatively, we need software techniques that can improve performance and energy efficiency on existing hardware.

In this paper, we determine that category-awareness introduces new opportunities for domain-specific approximation. Some categories are almost always larger than the other, e.g. train vs motorcycle, which results in different sensitivity to- ward approximation techniques. We explore the speed and accuracy trade-offs brought by two software-driven DSA optimization techniques: image size and the number of region proposals. Both of these techniques affect speed and accuracy [5]. Image size affects how much detail is covered within the receptive field of view, i.e., the region used to classify objects. Small images have more information crowded into a fixed size receptive field, and thus there are fewer regions that need examination, and this results in lesser accuracy but faster detection. Region proposals are the regions that are likely to contain objects. In object detection, algorithms that propose object regions often act as a filter that let regions with high probability of containing objects pass through them for further processing, e.g. classify where and what is the object within that region. Hence, reducing the number of region proposals reduces the total candidate regions that need examination and this results in lesser accuracy but faster detection.

We analyze category-aware domain-specific approximation (DSA) along two settings: static and dynamic. In the static case, we use one approximation configuration for each category. An approximation configuration refers to a tuple involving an image size and a region-proposal size. In the dynamic case, we select an approximation configuration per-image within a category. In both of the cases, our results indicate that category-awareness is promising in terms of speed improvement compared to the category-oblivious DSA dis- cussed in prior work [5]. In an oracle scheme, category-aware static and dynamic DSA achieve $3.7\times$ and $7.5\times$ speedup with- out accuracy degradation for a certain object category. Beyond our ideal oracle analysis, we also discuss the limitations and challenges of implementing both static DSA and dynamic DSA in a runtime and provide our insights toward harnessing the potential of both static and dynamic DSA. The runtime we design approximates the input frame dynamically by extracting useful features from previous frames and count on linear model to infer approximation. Our results show up to 51% and 211% speed improvement with 22% and 41% accuracy degradation

1: Department of Computer Science of National Chiao Tung University,
2: Department of Electrical and Computing Engineering, The University of Texas at Austin.

for Faster R-CNN and R-FCN, respectively.

In general, our findings indicate that future system designers of cognitive visual agents can improve the speed and energy-efficiency of them by using knowledge of what target object categories matter most. For example, an autonomous vehicle's system designer may want the vehicle's object detector(s) to work well on several crucial categories, such as pedestrians, bicyclists, and cars. One way to improve the speed of the system would be to try and find the image size that has the fastest speed but does not compromise accuracy across the three categories and deploy the system with the optimized image size. Alternatively, the designer could optimize the image size of the target categories and dynamically choose the corresponding image size when those objects show up in their perspective or field of view, leading to better performance.

In summary, our contributions are as follows:
- We investigate domain specific approximation (DSA) and characterize the effectiveness of category-awareness.
- We conduct limit study to understand the benefit of applying approximation in a per-frame manner with category-awareness, i.e. category-aware dynamic DSA.
- We present the challenges of harnessing domain specific approximation and our insight toward relaxing the challenges and provide our proof-of-concept runtime implementation.

The paper is organized as follows. Section 2 describes our experimental setup. Section 3 describes related state-of-the-art work and presents our insights and approach for DSA. Section 4 describes the challenges of DSA and our implementation and evaluation. Finally, Section 5 concludes the paper.

## II. EXPERIMENTAL SETUP

We conduct our analysis and optimization on two modern meta-architectures for deep learning-based object detection.

### A. Object Detectors

We study two state-of-the-art meta-architecture for deep learning-based object detection: Faster R-CNN [12] and R-FCN [2]. The feature extractors we use are the ones that come with the repository of the object detector, and the weights we use are the weights that were pre-trained on the PASCAL VOC dataset [3] by the original authors. Specifically, we use ZF network [16] for Faster R-CNN and ResNet-50 [4] for R-FCN.

We considered including both the SSD [9] and YOLO [11] network. However, the both of them are not amenable to input image scaling. The network demands a fixed size input. Both of them scale the image into a fixed size regardless of the original image size, and it performs poorly if the image resolution is large and contains a lot of small objects. As the camera resolutions continue to grow larger, we envision more and more applications will be based on more flexible meta-architectures, such as Faster R-CNN and R-FCN.

### B. Dataset and Metric

**Dataset** We picked 40 videos from ImageNet Object Detection from Videos (VID) task dataset for each of the categories we study and divide and into 20 training videos and 20 testing videos. Specifically, among the categories that overlapped with the PASCAL VOC dataset, we randomly pick eight categories for our analysis, which include the following: airplane, bus, car, dog, horse, motorcycle, sheep, and train. Notice that we pick categories that overlapped with the PASCAL VOC dataset, so that we can use the pre-trained models provided by authors without re-training.

**Metric** As in most of the object detection evaluation, we count on average precision or mean average precision to evaluate the accuracy of the object detector. We leverage the MS-COCO toolkit [8] and modify it to calculate per-image average precision for us to determine the optimal approximation configuration in a per-image granularity.

### C. Platform

**Hardware** We conduct all of our experiments on the SoC Tegra X1, which is a state-of-the-art embedded SoC that has a Maxwell GPU with 256 CUDA-core and a quad-core ARM A57 CPU. Unless stated as otherwise, the frequency of the CPU and GPU are kept constant at the highest speed, i.e. 1912.5 MHz for the CPU and 998.4 MHz for the GPU.

**Software** We use CUDA 7.0 and cuDNN 4 for GPGPU processing. We use the Caffe [6] deep learning framework to power all the algorithms we study.

## III. DOMAIN SPECIFIC APPROXIMATION (DSA)

In this section, we first describe related work in domain-specific approximation and the terms used throughout the paper. Then we analyze the effectiveness of category-aware static and dynamic DSA. Though we focus on performance, our results can be inferred toward improving energy-efficiency.

### A. State-of-the-Art

Approximation has been discussed in many recent works [7, 10, 14]. However, domain-specific approximation is still an emerging concept [13] and bound to specific domains, e.g. [1]. For object detection, the most relevant work is by Huang et al. [5], where the authors discuss trade-offs between speed and accuracy for different image sizes, the number of region proposals, and feature extractors in the object detectors.

Our work focuses on category-aware static and dynamic DSA. We suggest that category-awareness introduces more opportunities in DSA compared to category-oblivious DSA [5]. Additionally, we investigate the applicability of harnessing those opportunities with both static and dynamic DSA.

### B. Terminology and Definitions

For domain-specific approximation of object detection, we focus on image size and the number of region proposals.

**Downsampling** To set a proper baseline, we first scale every image in the dataset to a fixed height of 480 pixels. For approximating the image size, we evaluate 11 different downsampling levels, starting from 480 pixels. At each level, we downsample the image by 40 pixels, progressively scaling the image all the way down until we reach 80 pixels in height.

**Region Proposal** We choose 300 as the baseline. Both Faster R-CNN and R-FCN also use the same value by de- fault. Also,

we explore a different number of proposals (i.e., 300, 200, 100, 50, and 10) as a means to reduce processing requirements without impacting accuracy.

**ROI** We define an ROI (Region of Interest) as the bounding box surrounding the object of interest in an image.

**Approximation Configuration** We denote an approximation level as a configuration tuple. For example, (160, 50) denotes 160 in image height and 50 region proposals. We denote marginal approximation by leaving the other as a hyphen. For example, (360, -) means 360 in image height while using the baseline for the number of proposals, i.e. 300 in our study.

**Safe Approximation** Throughout the paper, we focus only on approximations that do not deteriorate the average precision. When we present results for the oracle approximation scheme, we select the approximation level that achieves equal or greater average precision compared to the baseline.

**Optimal Approximation** We define this as the fastest configuration (i.e., maximum FPS) while without accuracy degradation (i.e., still considered as a safe approximation).

### C. Category-Aware DSA Opportunity

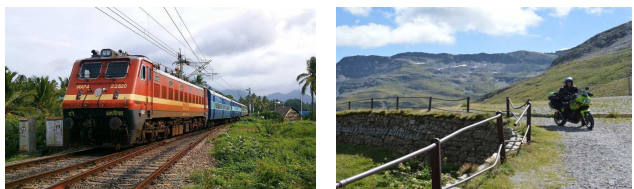

(a) Train    (b) Motorcycle

Fig. 1. Inter-category variation encourages category-aware domain specific approximation (DSA).

Some categories, such as the train shown in Fig 1 (a), are almost always larger in size than other categories, such as the motorcycle shown in Fig 1 (b). What this implies is that in terms of accuracy some categories almost always have lower sensitivity to approximation than others, which motivates us to investigate the benefits of bringing category-awareness to the domain-specific approximation techniques we study.

**Performance/Throughput** Before we analyze category's sensitivity to the domain specific approximation techniques, i.e. image scaling and proposal scaling, we first investigate the performance/throughput benefit brought by the techniques.

Fig. 2 shows an evaluation of how the number of frames per second (FPS) (i.e., performance/throughput) is affected by the two techniques. First, we find that for Faster R-CNN, due to the fully-connected layers applied to each of the region proposals, its speed is related to proposal scaling. For R-FCN, it is somewhat invariant to proposal scaling since the computation after ROI-pooling layer is small compared to the convolution layers before ROI-pooling. Second, both meta-architectures are sensitive to image scaling. Due to R-FCN's fully-convolutional design, it is more sensitive than Faster R-CNN. By jointly investigating both techniques, we find that Faster R-CNN can have better speed gain leveraging image scaling when the number of proposals is small. This is reasonable since fully-connected layers for each proposal imposes some computation

(a) Faster-RCNN    (b) R-FCN

Fig. 2. Throughput in FPS (shown within each cell) for object detection algorithms under two kinds of DSA techniques.

overhead, and it is invariant to image size. So, it's a competing effect between both techniques.

Similar approximation strategies have been explored in the past but in a more limited capacity. Huang et al. [5] investigate what knobs generally affect the speed and accuracy of the object detectors, while our analysis focuses more on image scaling and proposal scaling and is more comprehensive and insightful in the following manner. First, we analyze image sizes at a finer granularity, from 80 pixels to 480 pixels in height with a 40 pixels step. Second, our evaluation considers both techniques jointly, which introduces more approximation configurations with similar performance improvement for architectures that are sensitive to both techniques, i.e. Faster R-CNN. For example, Fig. 2a shows that approximation configurations (80, 300), (160, 50), and (240, 10) all share similar performance improvement, which enables the system to optimize for accuracy under the same performance improvement.

**Inter-Category Variance** Given the performance improvement motivation of the two DSA techniques, we next analyze the accuracy of the object detector for each category we study.

Fig. 3 indicates that Faster R-CNN is more sensitive to region proposal scaling compared to R-FCN. Moreover, R-FCN has a sharper boundary compared to Faster R-CNN, which suggests that Faster R-CNN is more robust when it comes to the two DSA techniques. Interestingly, different category has different sensitivity toward both DSA techniques. Airplane (Fig. 3a) has more "safe approximation" (see Sec. 3.2) configurations than train (Fig. 3b), which has more safe approximation configurations than motorcycle (Fig. 3c).

One of the reasons behind the variance is aligned with our intuition that the larger the object the smaller approximation can go without accuracy degradation. In our results, the average object size in pixels in the dataset we use for airplane, train, and motorcycle are 107k, 126k, and 38k, respectively.

However, inter-category variation is not solely due to the size of the objects but also the noise in the background, e.g. the background for the airplane class has less noise than others, and the fact that some categories are harder for the object detector than the others [15], which implies that the hard classes are more sensitive to information loss (approximation). The variation of the safe approximation across category encourages category-awareness for even faster system design.

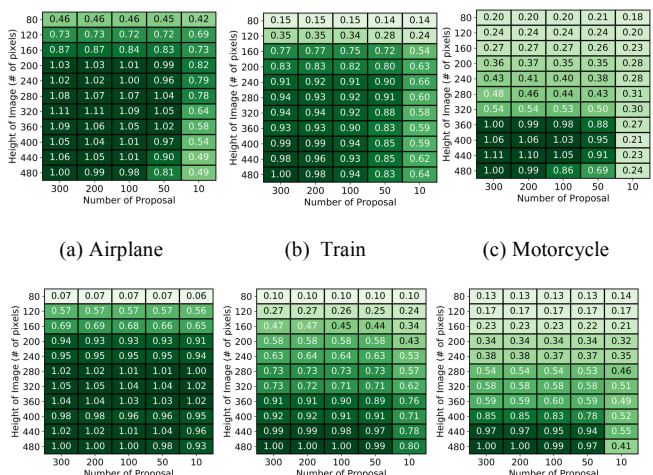

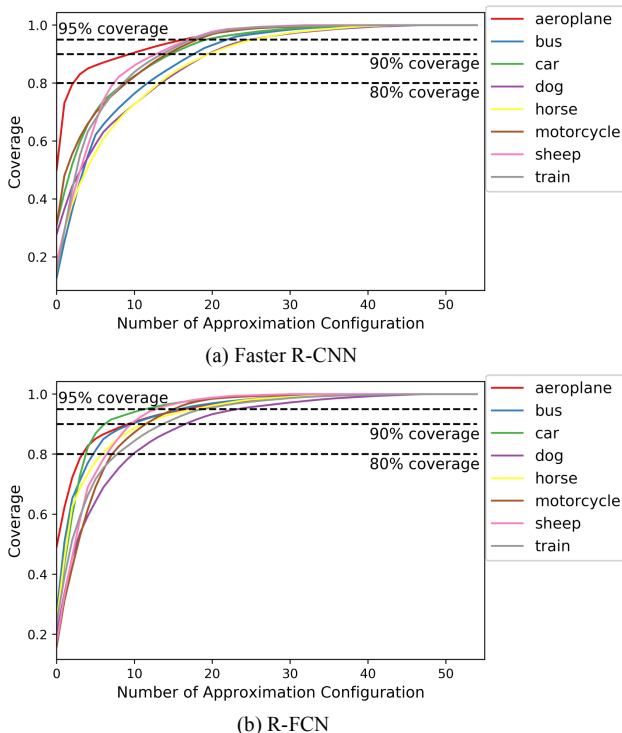

Fig. 3. Normalized average precision (shown inside each cell) of Faster R-CNN (top row) and R-FCN (bottom row) for three selected categories under two kinds of DSA techniques.

**Intra-Category Variance** Beyond the benefit introduced by category-awareness, we observe that there is even greater intra-category variance for domain-specific approximation techniques to exploit on a per-frame basis in an input stream.

In a dynamic input stream (e.g., video), the optimal approximation can be changing continuously. Objects in a dynamic input stream (e.g., video) are often constantly changing in perspective size (objects are sometimes near and large, and at other times small and far away). Fig. 4 shows a car passing by from left to right. Therefore, as time goes by the train gets larger in size, and so the level of safe approximation can be more aggressive over time. Therefore, dynamically picking an approximation configuration can sometimes be better than statically picking a one-time fixed DSA configuration.

To motivate "dynamic DSA," Fig. 5 shows the relationship between the portion of images of the training data that are approximated optimally and the corresponding number of approximation configurations considered for the eight categories that we mentioned in Sec. 2. For example, we need 18 different and unique approximation configurations as shown in Fig. 5a to optimally approximate 95% of the images of airplane from the training dataset.

The analysis hints us to whether a category is worthy of dynamic DSA. With static DSA, the system applies only one approximation configuration for the entirety of each category. However, for the most extreme case, such as the dog category in Fig. 5b, we require 25 approximation configurations to optimally approximate 95% of the dogs. This means that static DSA loses significant opportunity to improve performance. On the other hand, categories such as the sheep require the least number of approximation configurations to achieve 95%

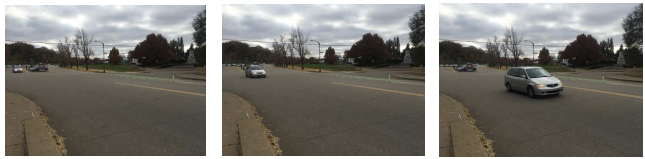

Fig. 4. Intra-category variation encourages dynamic DSA.

Fig. 5. The relationship between the portion of images of the training data that are approximated optimally, and the corresponding number of approximation configurations considered.

coverage, which may have better performance with static DSA.

In our analysis, dynamic DSA can speed up Faster R-CNN by up to 7.5x and R-FCN up to 7x without accuracy loss for some categories. So, if we can model the optimal approximation correctly with a simple model there is a large opportunity to speed up object detection-based systems significantly.

## IV. DSA CHALLENGES TOWARD IMPLEMENTATION

The limit study in Sec. 3 shows great promise in category-aware domain-specific approximation for object detectors. However, there are several challenges that remain to be resolved in order to harness the benefits. In this section, we will first illustrate the challenges for implementing category-aware DSA and then we provide our insights and implementation.

### A. Challenges

We pose two major challenges that we believe to be the key to unlock the potential for category-aware DSA and be useful.

**Prior Knowledge** Category-awareness exposes extra "slack" for domain-specific approximation techniques to exploit. However, without performing object detection in the first place it is unclear how the system can obtain the information regarding the category of the object to determine the approximation configuration. So, a key question is how to approximate the category information to begin with for starters?

**Dynamism Modeling** To further harness dynamic DSA, it is essential to understand what causes different optimal approximation configurations. It is critical to understand what the good features are to predict the optimal approximation. Moreover, to benefit from speed improvement the model for inferencing optimal approximation should be low-overhead.

*B. Implementation*

We implement a proof-of-concept dynamic DSA on streaming inputs, which is a common case in autonomous agents. In the streaming input case, there is temporal consistency, i.e. the difference between any two consecutive frames is small which we can leverage to overcome the aforementioned challenges.

**Prior Knowledge** With temporal consistency, we can resolve the first challenge approximately in the sense that we regard the category of the objects in the current frame to be the output of the object detector from the previous frames. Notice that by approximating the prior knowledge this way, we introduce two new problems. 1) How many in-coming frames can we assume to be the same as the current frame? 2) How much can we believe in the output of the object detector? For the first problem, we try one, three, and five frames and find three to be better than the others in the dataset we use. For the second problem, we count on the output probability, i.e. the output of the softmax layer to tell if we can trust the ROIs. We set a threshold to 0.6 in our implementation.

**Dynamism Modeling** We assume the approximation decision is correlated to the size and the number of objects in the frame. Intuitively, if the size of the object is large, we might be able to down-sample it more than when it is small. On the other hand, number of objects affect occlusion. If there are more objects in the frame, then it is more likely that there will be a lot of overlapping. Additionally, given a fix size input, the more objects there are in the frame, the smaller those objects can be. Hence, we build an ordered four polynomial regressor with the height and width of both the smallest and largest ROI in the frame and number of ROIs in the frame.

**Evaluation** We refer to our implementation as AutoFocus. We first evaluate the overhead introduced by AutoFocus. According to our measurements, on average the inferencing approximation configuration using the ordered four polynomial regressor takes 3.2 ms, which is only 0.6% of the original Faster R-CNN object detector's performance overhead.

We compare AutoFocus against static DSA with prior knowledge and category-oblivious DSA (prior work). To obtain results for the static DSA with prior knowledge, we obtain the optimal configuration for each category on the training set and use that configuration on the test set given we know the category in advance. For AutoFocus, we train the polynomial model on the training set and apply the model on the test set.

Fig. 6 shows our evaluation for category-oblivious DSA, static DSA and AutoFocus. We first focus on the benefits brought by static DSA compared to category-oblivious DSA. For categories like buses, cars, and motorcycles, static DSA improves the speed by 20% on average with merely 3% accuracy degradation, which exhibits the good part of static DSA. On the other hand, for categories like airplanes and dogs, there is a large margin in speed improvement, i.e. 60%, with relatively larger accuracy degradation, i.e. 22%, on average for Faster R-CNN. In general, static DSA introduces large performance benefit with little accuracy degradation. However,

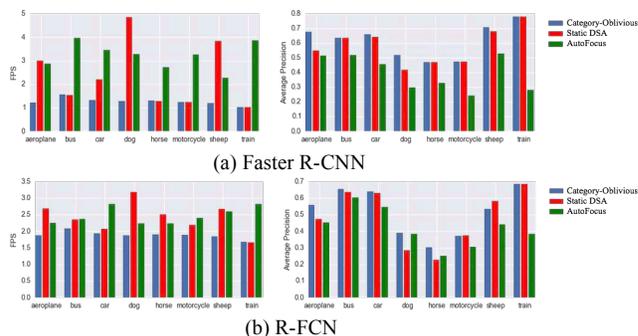

(a) Faster R-CNN

(b) R-FCN

Fig. 6. The performance and accuracy comparison among category-oblivious DSA, static DSA, and AutoFocus.

as stated previously, it requires prior category knowledge.

For AutoFocus, there is a general trend across categories that AutoFocus brings larger speed-up improvement but with relatively larger accuracy degradation compared to static DSA. It introduces 51% and 211% speed improvement with 22% (0.09) and 41% (0.22) accuracy (average precision) degradation on average for Faster R-CNN and R-FCN, respectively. The primary reason for the accuracy degradation in our run-time's inefficiency. It is failing to accurately model the intra-category dynamism. As mentioned in Sec. 3, Faster R-CNN is more robust to both DSA techniques, which results in the accuracy degradation difference. However, if we compare speed-up and accuracy degradation equally, our runtime, on average, still outperforms static DSA with prior knowledge.

## V. CONCLUSION

In this work, we conduct a limit study on category-aware static and dynamic domain-specific approximation and present the challenges and our insights toward them. Our result shows that category-aware domain-specific approximation opens new opportunity for better speed or energy-efficiency and takes a step toward its realization. However, to harness the benefit, we think more research effort should be placed to better addressed the challenges. In the long-run, as visual systems become increasingly more intelligent, we believe DSA will offer significant improvements to system designers that are tasked with the trade-offs between performance, power and accuracy.

**Ting-Wu Chin** is a Ph.D. student working with Dr. Diana Marculescu in the Department of Electrical and Computer Engineering at Carnegie Mellon University. His research interests broadly cover machine learning, AI systems, energy-aware computing, and mobile cloud computing. He received both his MS and BS in Computer Science from National Chiao Tung University at Hsinchu, Taiwan. Contact him at tingwuc@andrew.cmu.edu.

**Chai-Lin Yu** received the BS degree and the MS degree in Computer Science from National Chiao Tung University in 2015 and 2017, respectively. He is currently a software engineer at MediaTek inc., Taiwan. His research interests include high-performance heterogeneous computing, automatic performance tuning, and performance modeling. Contact him at tony1223yu.cs00@g2.nctu.edu.tw.

**Matthew Halpern** Matthew Halpern is a PhD student in the Department of Electrical and Computer Engineering at the University of Texas at Austin. His research interests include runtime systems and computer architecture for mobile computing. Halpern has a BS in Electrical and Computer engineering from the University of Texas at Austin. Contact him at matthalp@utexas.edu.

**Hasan Genc** is an undergraduate student in the Department of Electrical and Computer Engineering at the University of Texas at Austin under the supervision of Dr. Vijay Janapa Reddi. His research interests include computer architecture, operating systems, and software-hardware co-design. Contact him at hngenc@utexas.edu.

**Shiao-Li (Charles) Tsao** earned his PhD degree in engineering science from National Cheng Kung University, Taiwan in 1999. His research interests include energy-aware computing, embedded software and system, and mobile communication and wireless network. He was a visiting professor at Dept. of Computer Science, ETH Zurich, Switzerland, in the summer of 2010 and 2011, and from 2012 to 2013. From 1999 to 2003, Dr. Tsao joined Computers and Communications Research Labs (CCL) of Industrial Technology Research Institute (ITRI) as a researcher and a section manager. Dr. Tsao is currently a professor of Dept. of Computer Science and director of Institute of Computer Science and Engineering of National Chiao Tung University. Prof. Tsao has published more than 110 international journal and conference papers, and has held or applied 24 US patents. Prof. Tsao received the Young Engineer Award from the Chinese Institute of Electrical Engineering in 2007, Outstanding Teaching Award of National Chiao Tung University, K. T. Li Outstanding Young Scholar Award from ACM Taipei/Taiwan chapter in 2008, and 2013 Award for Excellent Contributions in Technology Transfer from Ministry of Science and Technology. Contact him at sltsao@cs.nctu.edu.tw.

**Vijay Janapa Reddi** is as an Associate Professor in the Department of Electrical and Computer Engineering at the University of Texas at Austin. His research interests include architecture and software design to enhance system performance, user experience, energy efficiency and reliability for consumer devices and autonomous systems. He has a PhD in Computer Science from Harvard University, a MS from the University of Colorado at Boulder and a BS from Santa Clara University. Contact him at vj@ece.utexas.edu.